\documentclass[journal]{IEEEtran}

\usepackage{times}
\usepackage{graphicx}
\usepackage{epsfig}
\usepackage{epstopdf}
\usepackage{CJK}
\usepackage{indentfirst}
\usepackage{amsmath}
\usepackage{amssymb}
\usepackage{mathrsfs}
\usepackage{subfigure}
\usepackage{booktabs}
\usepackage[numbers,sort&compress]{natbib}
\usepackage{flushend}
\usepackage{amsthm}
\usepackage{amssymb}
\usepackage[ruled]{algorithm2e}
\usepackage{makecell}
\usepackage{booktabs}
\usepackage{soul}
\usepackage{multirow}
\usepackage{bm}

\begin{document}

\title{A Study on Evaluation Standard for Automatic Crack Detection Regard the Random Fractal}

\author{Hongyu Li,
        Jihe Wang,
        Yu Zhang$^*$,
        Zirui Wang,
        and Tiejun Wang$^*$
\thanks{H. Li, Z. Wang, and T. Wang are with School of Aerospace and State Key Laboratory for Strength and Vibration of Mechanical Structures, Xi'an Jiaotong University, Xi'an, Shaanxi, 710049, China (e-mail: lihongyu\_0@163.com; wangzirui@stu.xjtu.edu.cn; wangtj@xjtu.edu.cn).}
\thanks{J. Wang is with School of Computer Science, Northwestern Polytechnical University, Xi'an, Shaanxi, 710068, China, (e-mail: wangjihe@nwpu.edu.cn).}
\thanks{Y. Zhang is with School of Computer Science, Shaanxi Normal University, Xi'an, Shaanxi, 710119, China, (e-mail: zhangyu82@snnu.edu.cn).}
\thanks{* denotes the corresponding authors.}
}
\maketitle

\begin{abstract}
A reasonable evaluation standard underlies construction of effective \textit{deep learning} models.
However, we find in experiments that the automatic crack detectors based on \textit{deep learning} are obviously underestimated by the widely used \textit{mean Average Precision} (mAP) standard.
This paper presents a study on the evaluation standard.
It is clarified that the random fractal of crack disables the mAP standard, because the strict box matching in mAP calculation is unreasonable for the fractal feature.
As a solution, a fractal-available evaluation standard named CovEval is proposed to correct the underestimation in crack detection.
In CovEval, a different matching process based on the idea of covering box matching is adopted for this issue.
In detail, \textit{Cover Area rate} (CAr) is designed as a covering overlap, and a multi-match strategy is employed to release the one-to-one matching restriction in mAP.
\textit{Extended Recall} (XR), \textit{Extended  Precision} (XP) and \textit{Extended F-score} ($F_{ext}$) are defined for scoring the crack detectors.
In experiments using several common frameworks for object detection, models get much higher scores in crack detection according to CovEval, which matches better with the visual performance.
Moreover, based on faster R-CNN framework, we present a case study to optimize a crack detector based on CovEval standard.
Recall (XR) of our best model achieves an industrial-level at 95.8, which implies that with reasonable standard for evaluation, the methods for object detection are with great potential for automatic industrial inspection.
\end{abstract}


\IEEEpeerreviewmaketitle

\section{Introduction}

\IEEEPARstart{C}{racks} are dangerous defects in structure and machinery.
Any structural failure caused by cracks on crucial components could lead to catastrophic accidents.
To ensure the safety of in-service equipments, regular detection for these potential dangers is required at scenarios such as pressure vessels, steam turbines in energy industry, and aero-engines, landing gears in aviation industry \cite{Aldea2018Robust,VARNEY2012Crack}.
Visual inspection is the most fundamental approach.
Plenty of time and costs are spent on traditional manual detection~\cite{2010Structural}, while the quality is strongly influenced by concentration and experience of human inspectors.
Hence, to achieve efficient visual inspection, it is a significant challenge in industry to detect cracks automatically from images.

There are various methods for automatic crack detection.
Specifically, manual designed methods~\cite{Chisholm2020FPGA,Oliveira2013Automatic,Shi2016Automatic} could not achieve good performance in complex environments.
Approaches based on \textit{deep learning} and \textit{Convolutional Neural Networks} (CNN) are effective in processing the images~\cite{Dong2019PGANet,Gao2020Generative}.
For crack detection, these methods are more robust~\cite{Chen2018NB-CNN,Du2020Intelligent,Feng2017Deep,Cha2017Deep}, they can not locate the targets from large images.
These methods need to scan all region of the image for the defects, which is inefficient for application.
Methods for \textit{object detection}, which is remarkable and flourish in \textit{deep learning}~\cite{Su2020Deep,Masood2020Automated}, are robust, efficient and more practical for crack detection.
By region-based strategies~\cite{Girshick2015Fast,Ren2015Faster,Cai2017Cascade}, these methods can mark the targets with rectangular boxes directly.
Cha Y et al.~\cite{Cha2017Autonomous} successfully applied these methods for defect detection (including the cracks) in civil infrastructures.
Actually, the practical and great potential methods for \textit{object detection} should be better promoted and developed in industrial defect detection.

\begin{figure}[t]
\centering
\includegraphics[width=0.98\linewidth]{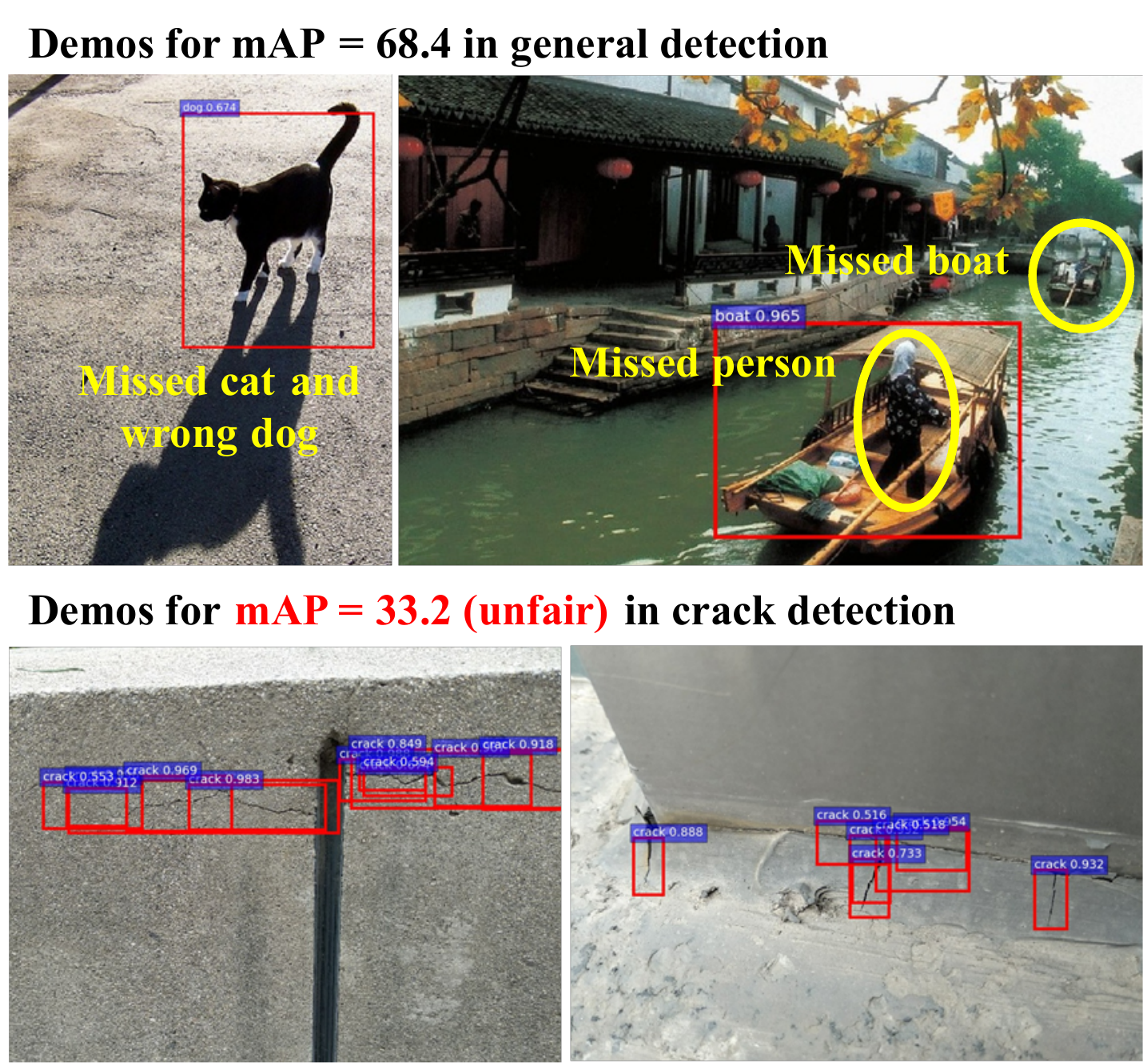}
   \caption{Conflict between poor \textit{mean Average Precision} (mAP) score and fine visual performance in crack detection. Demo images and their AP scores of the general objects (left) and cracks (right) are presented. According to the visual performance, the crack detector should not be such low scored (33.2).}
   \vspace{-5pt}
\label{fig:motivation}
\end{figure}

Reasonable standard for scoring the models is the prerequisite to construct powerful methods for crack detection.
Conventional standards in machine learning~\cite{Fawcett2006An,Hand2009AUC} are not applicable for the emerging methods for \textit{object detection}, because the ability in positioning and classifying should be evaluated jointly for models in this kind.
Current widely used standard for \textit{object detection} is the \textit{mean Average Precision} (mAP), which is designed and used in some international challenges and benchmark datasets~\cite{Everingham2006PASCALvoc,lin2014microsoft}.
This standard is designed for general object detection in daily life, and it calculates the final score by matching the detected boxes and \textit{Ground Truth} (GT) boxes for validation.
However, there are controversies about this standard in general object detection~\cite{Chavali2016Object,Hosang2016What}, that the mAP scores could not indicate the visual performance of the detection reasonably.
It could be more serious in special scenarios like the crack detection in industry.
In the few years since methods for \textit{object detection} appeared, discussions about the standard are rare.
Only a few studies extended this evaluation standard for object detection in autonomous driving scenarios~\cite{abs-1903-07840}.

Serious underestimation of the automatic crack detectors is observed in our practices that scores of the crack detector are really poor, while the visualized images of crack detection present competitive performance (Fig.~\ref{fig:motivation}).
This phenomenon implies that the crack is a kind of special object, and mAP standard is not suitable for evaluating the crack detectors.

This underestimation is attributed to a unique graphic characteristic of crack, the random fractal.
By theoretical analyses, we reveal that random fractal of crack lead the models for \textit{object detection} to present a special box marking mode.
However, this mode does not match the requirements of the mAP standard.
In detail, we find two issues in the box matching process, the \textit{scale inconsistency problem} (box scale mismatch) and \textit{non-correspondence problem} (the mismatch between box groups).
Based on the idea of covering box matching, we propose CovEval standard.
These problems discovered are addressed by new overlap called \textit{Cover Area rate} (CAr) and a multi-match strategy on boxes in CovEval, then, the boxes (the detected boxes and GT boxes for validation) can be matched reasonably.
Result of the experiments fully supports our analysis about crack and shows the effectiveness of the novel CovEval standard proposed.
Several popular generic frameworks for \textit{object detection} are evaluated by mAP and CovEval respectively.
Scores of crack is greatly increased via CovEval standard compared to general objects, which corresponds better with the observed visual performance.

Our main contributions are summarized as follow:
\begin{enumerate}
\item We revealed that the crack detectors based on \textit{deep learning} are seriously underestimated by the mAP standard due to the failed box matching.
\item Random fractal of crack is discovered, and we revealed that the special marking mode of crack caused by random fractal does not match the requirements of the mAP standard.
\item A new evaluation standard with more rational box matching for cracks, CovEval, is proposed to provide fair evaluation for the crack detectors.
\item With CovEval, an outstanding crack detector is successfully trained and optimized.
    It suggests that methods for \textit{object detection} are powerful in industrial inspection and CovEval removes the obstacle in evaluating the models.
\end{enumerate}

\section{Problem Statement: the random fractal of crack disables the mAP standard}
\label{Analysis of the underestimation}

Crack is random fractal, which is considered as the origin of the underestimation.
This section presents an in-depth theoretical study about the random fractal of crack, the marking mode of boxes, and the failure reason of mAP.
These analyses reveal mechanism of the underestimation.

\subsection{Preliminary: fractal theory and Rf object}

\begin{figure}[t]
\vspace{-5pt}
\centering
\subfigure[regular fractal transform $\mathscr{F}_K$]
    {
    \label{regular_fractal}
    \includegraphics[width=0.45\linewidth]{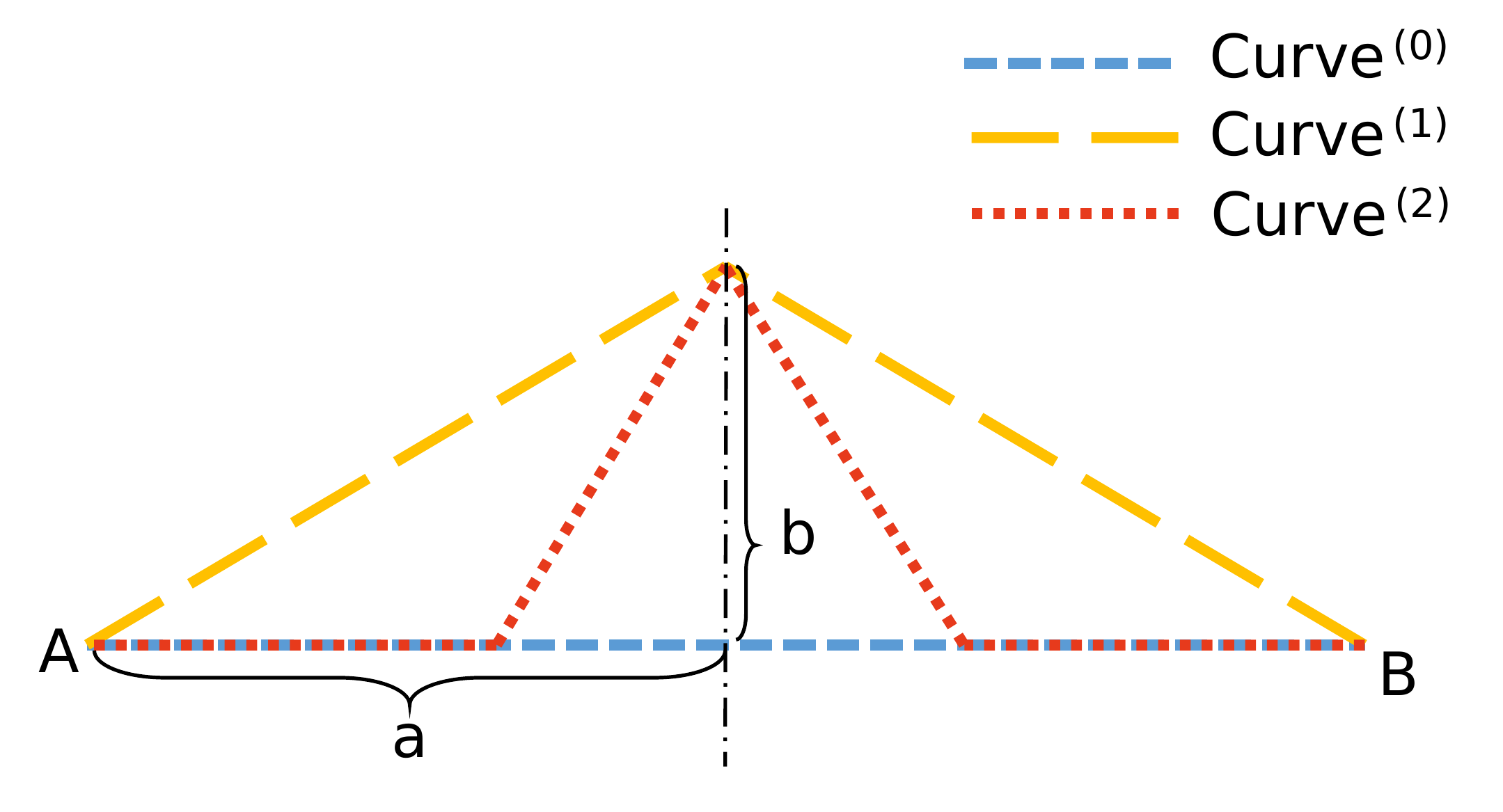}
    }
\subfigure[random fractal transform $\mathscr{F}_R$]
    {
    \label{random_fractal}
    \includegraphics[width=0.45\linewidth]{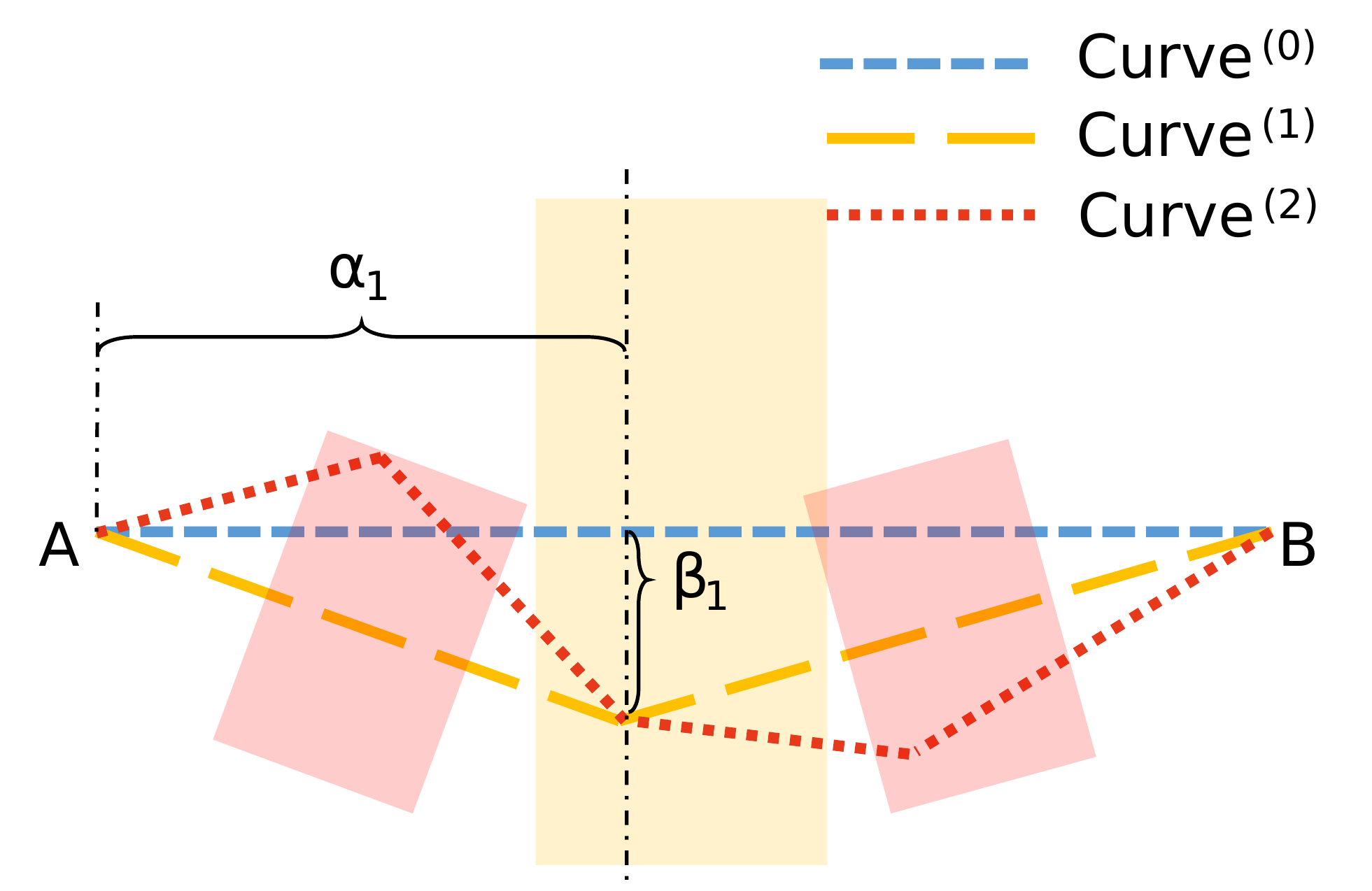}
    }
\subfigure[Fractal curve generated by $\mathscr{F}_K$]
    {
    \label{fractal_curve}
    \includegraphics[width=0.7\linewidth]{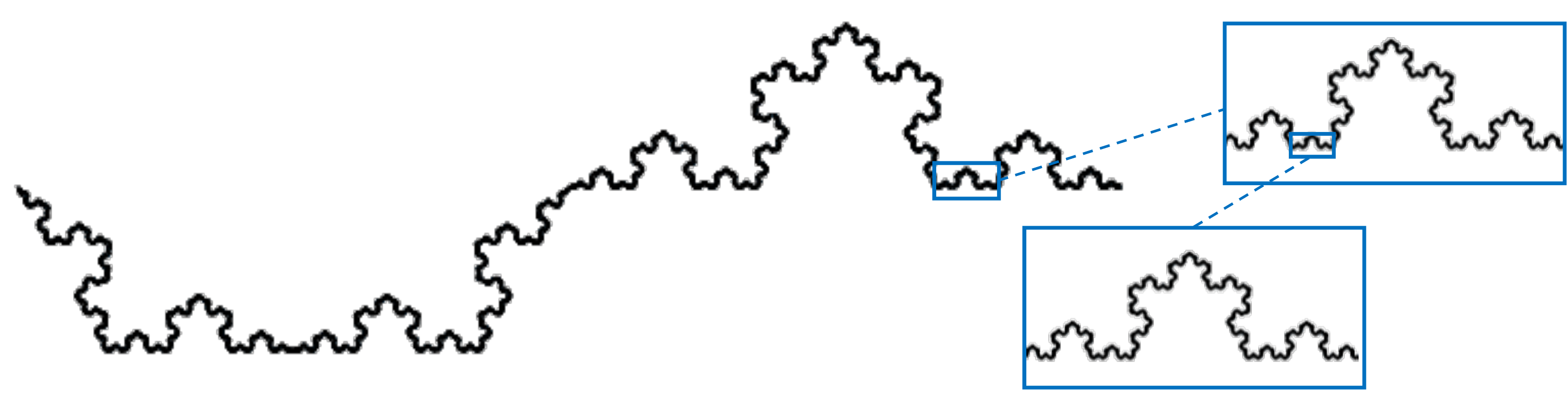}
    }
\subfigure[Random fractal curve generated by $\mathscr{F}_R$]
    {
    \label{random_fractal_curve}
    \includegraphics[width=0.7\linewidth]{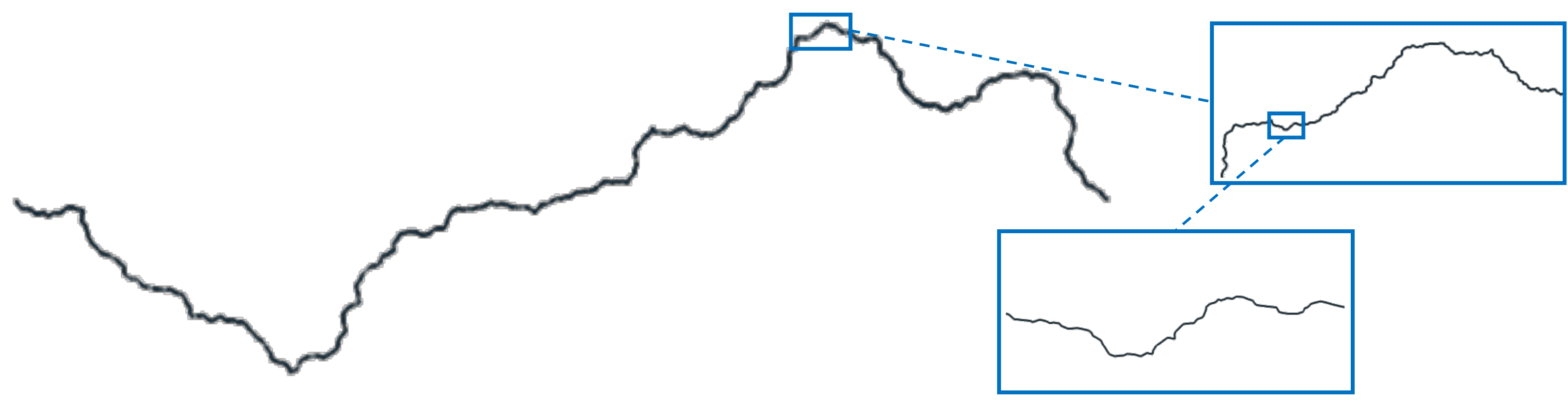}
    }

\caption{Fractal transform method $\mathscr{F}_K(\overline{AB},\{a,b\})$ and random fractal transform method $\mathscr{F}_R(\overline{AB},\{\alpha,\beta\})$ with their generated curves.
In this case, $a,b$ are constants; $\alpha,\beta$ are independent random variables that have $\alpha \sim U(\zeta_t , \zeta_b)$ and $\beta \sim U(\xi_t , \xi_b)$, where $\zeta_t$, $\zeta_b$, $\xi_t$ and $\xi_b$ are constants to determine the distribution range of the new nodes (shown in figure (b) with the colored areas).}
\label{fig:Transform}
\end{figure}

\begin{figure}[htbp]
\centering
\includegraphics[width=0.9\linewidth]{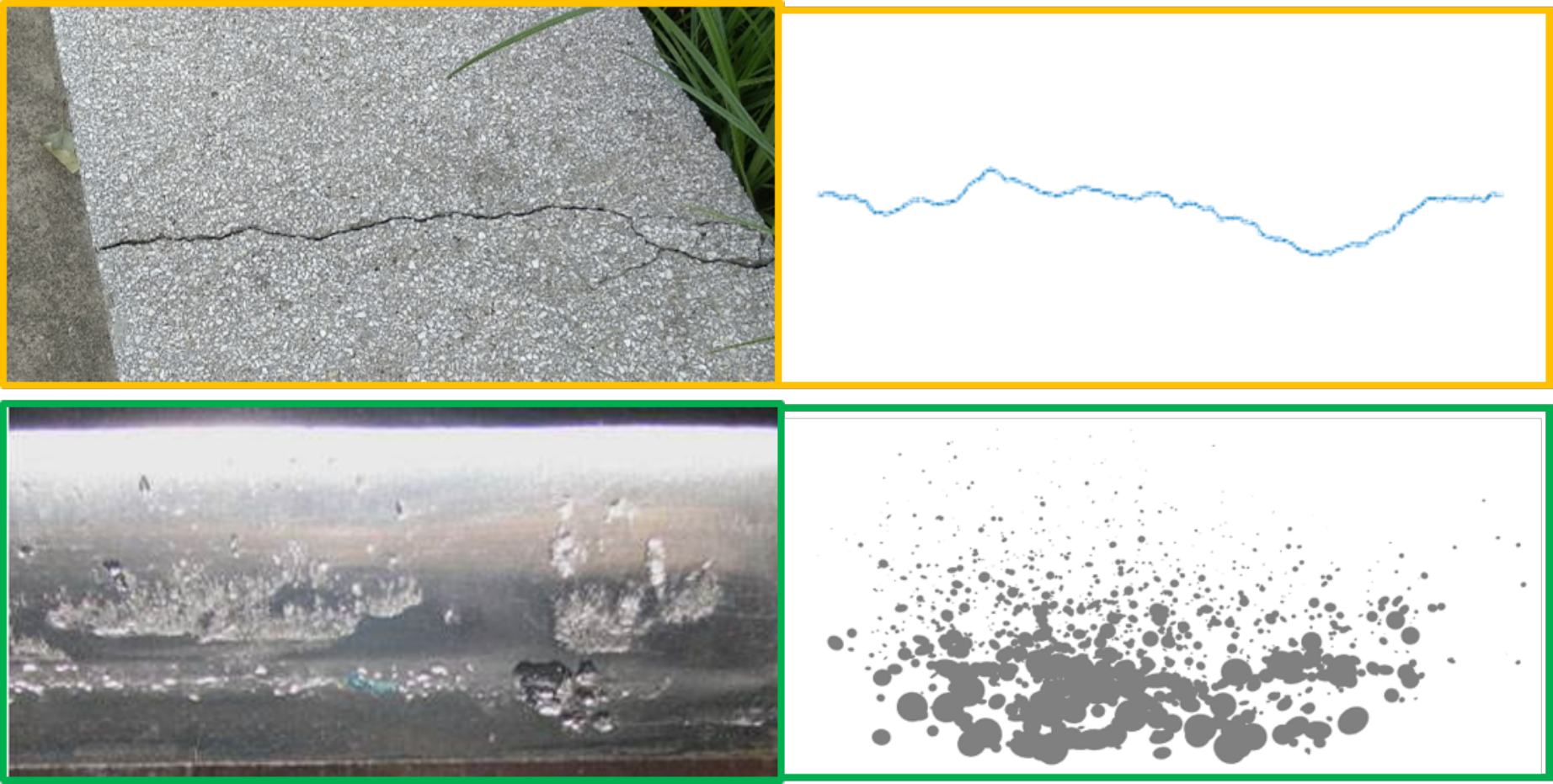}
\caption{Defect images and their simulated graphics. The geometries on the right are generated by program using random fractal method. Note that region defects like the pitting can also be considered as \textit{Random-fractal object} (Rf object).}
\label{fig:Fractal}
\vspace{-10pt}
\end{figure}

Fractal graph is ubiquitous~\cite{Mandelbrot1967How}.
A mathematical description is given here about the process to generate fractal curves:
\begin{align}
\label{iteration}
Curve^{(n+1)} = \mathscr{F}(\{Seg_k|Seg_k \subseteq Curve^{(n)}\}, \{\theta_i\})
\end{align}
\begin{align}
\label{fractal_definition}
Curve_{fractal} = {\lim_{n \to \infty}} \mathscr{F}^{(n)}(Curve^{(0)}, \{\theta_i\})
\end{align}
$\mathscr{F}(Inputs, Parameters)$ is defined as a transform method for generating fractal curves iteratively.
With Eq.~(\ref{iteration}, \ref{fractal_definition}), all segments \{$Seg_k$\} of the input $Curve^{(n)}$ are transformed iteratively with a certain method $\mathscr{F}$ parameterized by $\{\theta_i\}$.
Different from the ordinary fractal, the random fractal curves~\cite{Saupe1988RandFractal} are parameterized by the random variables $\{\Theta_i\}$.
For instance, Fig.~\ref{fig:Transform} shows a fractal transform method $\mathscr{F}_K$, a random fractal transform method $\mathscr{F}_R$, and two fractal curves generated respectively following these two methods. The general one is similar with Koch curve~\cite{Bannon1991Fractals}, and the random fractal curve is similar in graphic with the cracks (Fig.~\ref{fig:Fractal}). On this basis, we regard crack as object with random fractal features, the \textit{Random-fractal object} (Rf object).

\begin{figure*}[t]
\centering
\vspace{-10pt}
\includegraphics[width=0.96\linewidth]{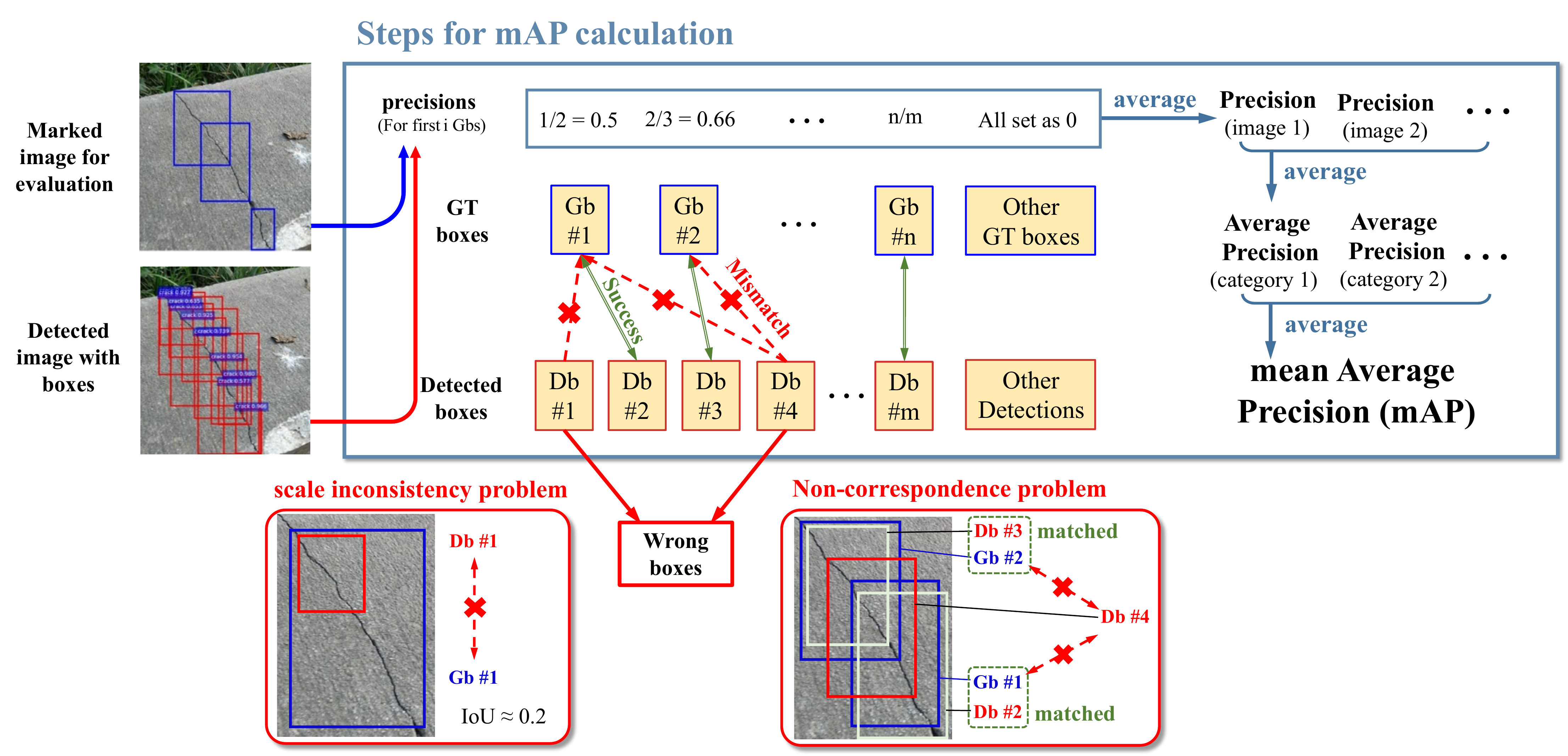}
   \vspace{5pt}
   \caption{Calculating process of mAP. Two issues in box matching are shown in the red boxes.
   The sorted detected boxes (Dbs) and GT boxes (Gbs) are matched according to their overlap.
   All detected boxes in the demo pictures are visually correct, but Db$\sharp1$ (It is failed to get match with Gb$\sharp1$ due to the scale difference) and Db$\sharp4$ (It is failed to get match with Gb$\sharp1$ or Gb$\sharp2$ due to the one-to-one corresponding criterion) are finally misjudged into wrong boxes.
   These two cases are examples of the \textit{scale inconsistency problem} and \textit{Non-correspondence problem} respectively.}
\label{fig:mAP}
\vspace{-5pt}
\end{figure*}

\subsection{Box marking mode for cracks (Rf objects)}
\label{theory}

Methods for \textit{object detection} identify the targets by rectangular boxes according to the included graphic features.
From demos in Fig.~\ref{fig:motivation}, boxes for marking cracks shows a repeating continuous mode.
By the following analysis, the mechanism of this special box marking mode is revealed based on the feature of Rf objects.
It starts with a theorem to clearly define the equivalent features of fractal curves.

\newtheorem{thm}{\bf Theorem}
\begin{thm}\label{thm1}
\textbf{All-scale equivalence of the fractal.} For a fractal curve $C_{f}$, any non-infinitesimal continuous subpart $C_s \subseteq C_{f}$ is composed by a series of curves $\{C_{i}\}$ that all the $C_{i}$ have equivalent graphic feature with $C_{f}$.

(the proof refers to Appendix~\ref{proof1})
\end{thm}

\begin{figure}[hb]
\centering
\vspace{-5pt}
\includegraphics[scale=0.55,width=1\linewidth]{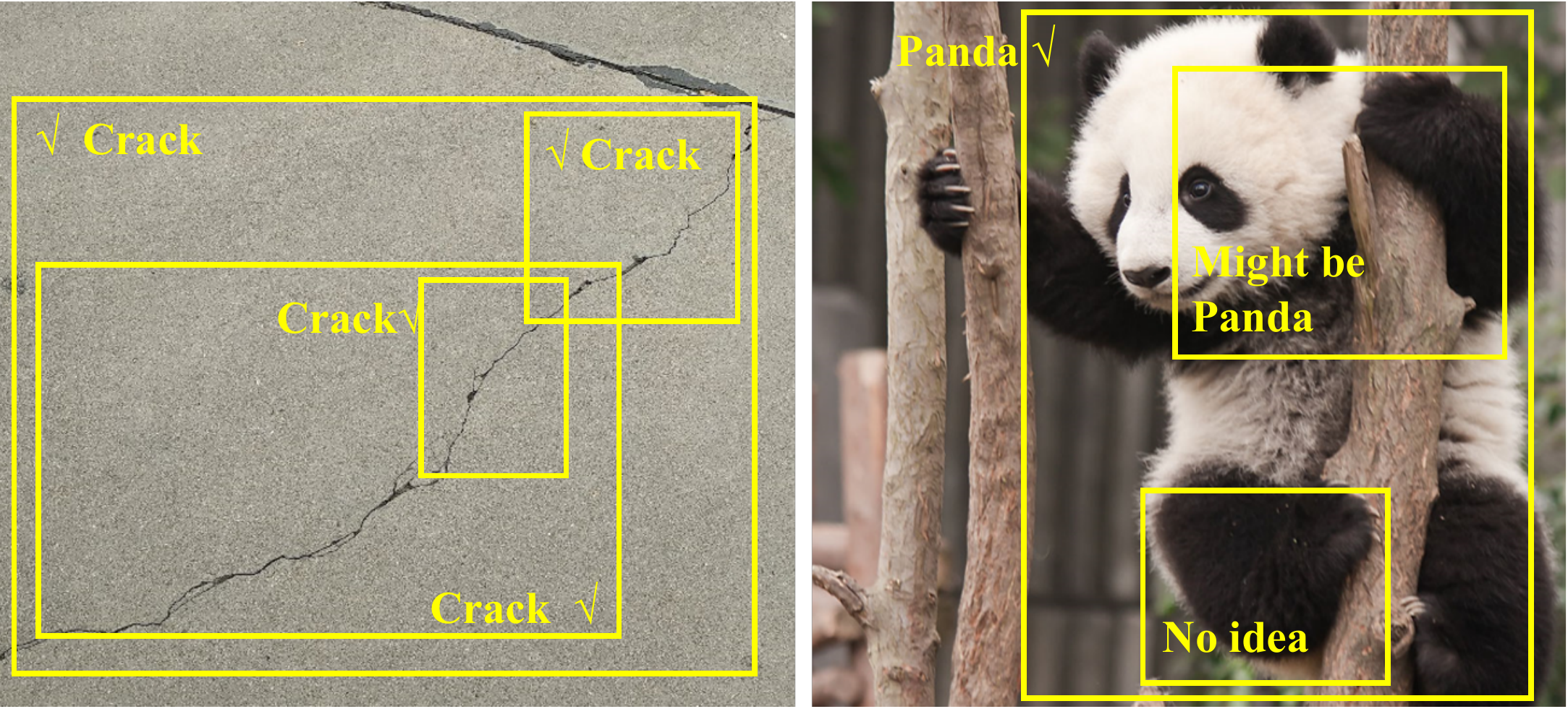}
   \caption{A comparison of crack (Rf object) and panda (general object). Boxes in the images are hypothetical detected boxes or GT boxes for validation. Every box has a verdict to determine whether it includes adequate graphic features to describe an object.}
\label{fig:Comarison}
\vspace{-5pt}
\end{figure}

Theorem~\ref{thm1} means that for any box including part of the crack, it includes full feature of crack, thus, marking cracks with repeating continuous boxes like demos in Fig.~\ref{fig:motivation} is reasonable. It is a special characteristic of crack which is called \textit{multi-scale segmentable}, which lead the models to give out repeating continuous boxes. Fig.~\ref{fig:Comarison} shows this characteristic of crack by example.

\subsection{Issues in evaluation using the mAP}

Methods for \textit{object detection} are scored by matching the detected boxes and GT boxes. Prior to analysing the issues, we give a brief review of the mAP standard as follow.

There are two main steps to calculate the mAP score: 1) calculate overlaps between the boxes, 2) calculate the score by matching the boxes according to the overlaps. Firstly, the overlap which represents correspondence of two boxes, is calculated by IoU:
\begin{align}
\label{fun:IoU}
IoU(G, D)=\frac{S_{G \bigcap D}}{S_{G \bigcup D}}
\end{align}
$S_{G \bigcap D}$ and $S_{G \bigcup D}$ represent area of intersection and union of box $G$ and $D$.
According to a threshold of overlap, the two boxes can be matched or not.
As shown in Fig.\ref{fig:mAP}, the detected boxes are listed by their confidence, \textit{precision} ($p$) of the first $n$ valid detected objects can be calculated: $p(n)=\frac{n}{m}$.
$m$ is the ranking of detected box which matches to the $n^{\textup{th}}$ GT box. Then, \textit{Precision} ($P$) is the average of $p$ for all the GT boxes (set $p=0$ for undetected ones).
Continuously, by averaging $P$ of images for one category or all categories, we can get \textit{Average Precision} (AP) score for one category or the mAP score for all categories respectively.

In case of crack detection, the long cracks are always marked with groups of repeating continuous boxes instead of individual ones, and boxes in the groups are arbitrary.
In detail, there are two features about the boxes.
1) The boxes are in different size.
2) The amount and position of boxes are uncertain, and they are repetitive.
These features impact the two main steps of mAP calculation respectively.
Firstly, marking Rf object with large or small boxes is both reasonable in practice, but boxes with different size can never get high overlap via IoU.
This conflict is referred as the \textit{scale inconsistency problem}.
Secondly, the number of boxes is arbitrary, and the boxes are dispersed over a considerable range.
Two groups of arbitrary boxes around one crack are almost impossible to achieve a good one-to-one match.
This conflict is referred as the  \textit{non-correspondence problem}.
By example, Fig.\ref{fig:mAP} includes illustrations about these two issues.

~

In brief, by revealing the mechanism of the mAP's failure in crack detection, \textit{how the random fractal causes these issues in evaluation}, is answered.
Specifically, the crack detectors are underestimated by the unreasonable box matching criterion in mAP standard.
We point out two issues in box matching, 1) the \textit{scale inconsistency problem}, 2) the  \textit{non-correspondence problem}.
Therefore, constructing reasonable box matching rules against this two issues is the key to correct this underestimation.

\section{CovEval: A fractal-available standard for crack detection}
\label{Proposed method}

We propose a novel evaluation standard called CovEval\footnote{The CovEval is named by the shorthand of \textit{Covering Evaluation}.} for scoring the crack detectors.
Covering is the core idea of the CovEval, which is the key to address the \textit{scale inconsistency problem} and \textit{non-correspondence problem}.
Two strategies based on the idea of covering are employed for matching the box groups of cracks.

\subsection{Scale-unrelated covering overlap: Cover Area rate}
\label{CAr}

When matching the boxes, overlap between every two boxes is the criterion.
The \textit{scale inconsistency problem} appears in the overlap calculation, that the boxes in different sizes can never be matched according to traditional IoU in Eq.~(\ref{fun:IoU}).
We construct a new scale-unrelated overlap called \textit{Cover Area rate} (CAr) as:
\begin{align}
CAr(G,D)=\frac{S_{G \bigcap D}}{min(S_G,S_D)}
\label{fun:CAr}
\end{align}
$D$ is a detected box; $G$ is a GT box; $S_G$ and $S_D$ are areas of $G$ and $D$ respectively; $S_{G\bigcap D}$ represents the intersection area of $G$ and $D$. Our implementation to calculate the CAr is shown in Algorithm~\ref{algorithm1}.

\begin{algorithm}[ht]
\caption{Calculate CAr for box $B_A$ and $B_B$}
\KwIn{ locating parameters of box $B_A$ and $B_B$:
$\{x_{A1}, y_{A1}, x_{A2}, y_{A2}\}$,
$\{x_{B1}, y_{B1}, x_{B2}, y_{B2}\}$}

assert: $x_{1} < x_{2}$ and $y_{1} < y_{2}$ for both $B_A$ and $B_B$\;

\KwOut{$CAr(B_A,B_B)$ }

\eIf{$x_{A1} > x_{B2}$~or~$y_{A1} > y_{B2}$~or~$x_{B1} > x_{A2}$~or~$y_{B1} > y_{A2}$}
{
    return 0\;
}
{
    compute box area $Area_A$ and $Area_B$\;
    sort all $x$ as $\{x_a,x_b,x_c,x_d\}$\;
    sort all $y$ as $\{y_a,y_b,y_c,y_d\}$\;
    $Area_U = (x_c-x_b)(y_c-y_b)$\;
    $CAr(B_A,B_B)=Area_U~/~min(Area_A, Area_B)$\;
}
return $CAr(B_A,B_B)$\;
\label{algorithm1}
\end{algorithm}

\begin{figure}[t]
\centering
\subfigure[Scale normalization]
    {
    \label{fig:normed-overlap}
    \includegraphics[width=0.43\linewidth]{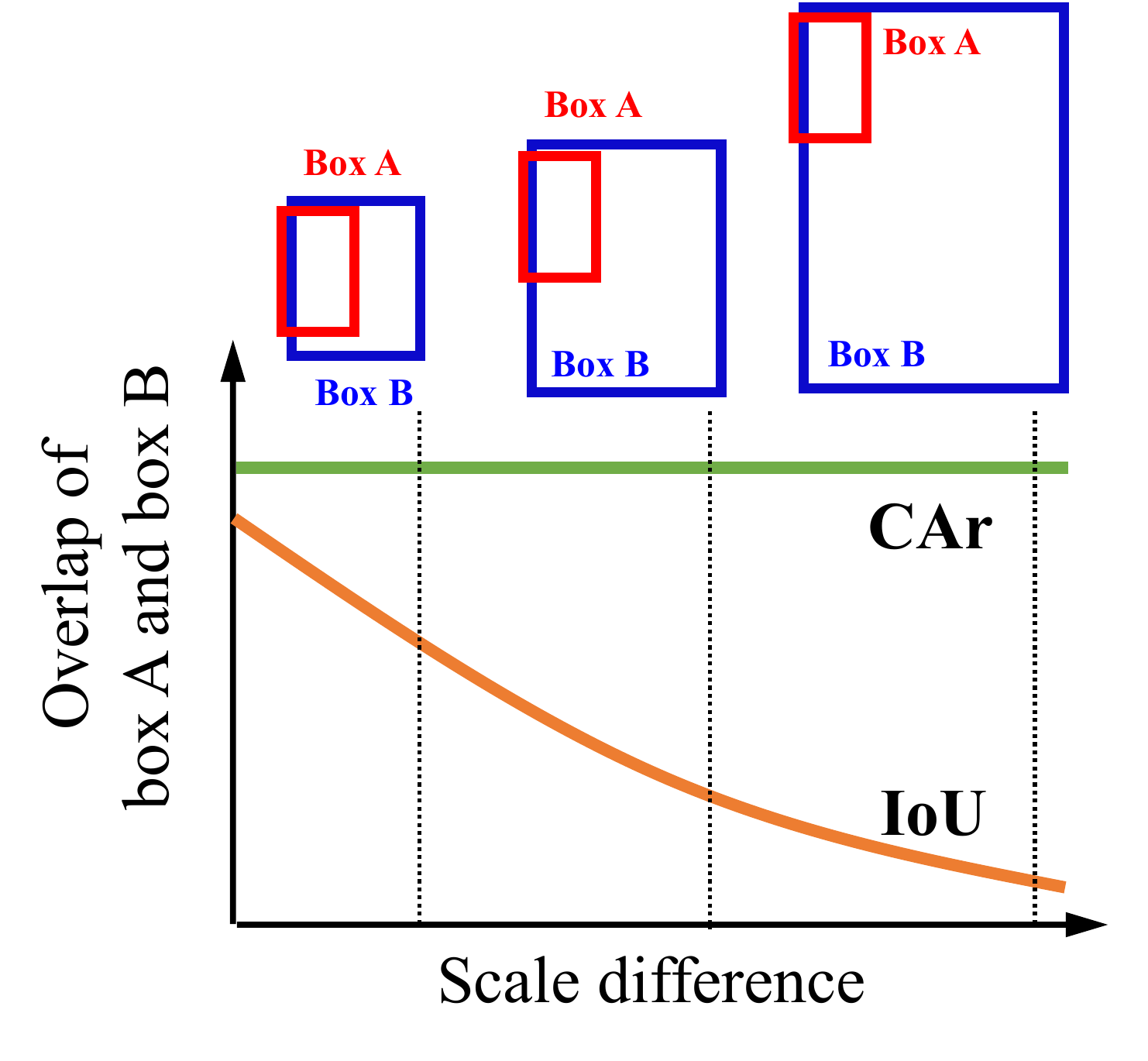}
    }
\subfigure[Multi-matching process]
    {
    \label{fig:multi-corresponding}
    \includegraphics[width=0.50\linewidth]{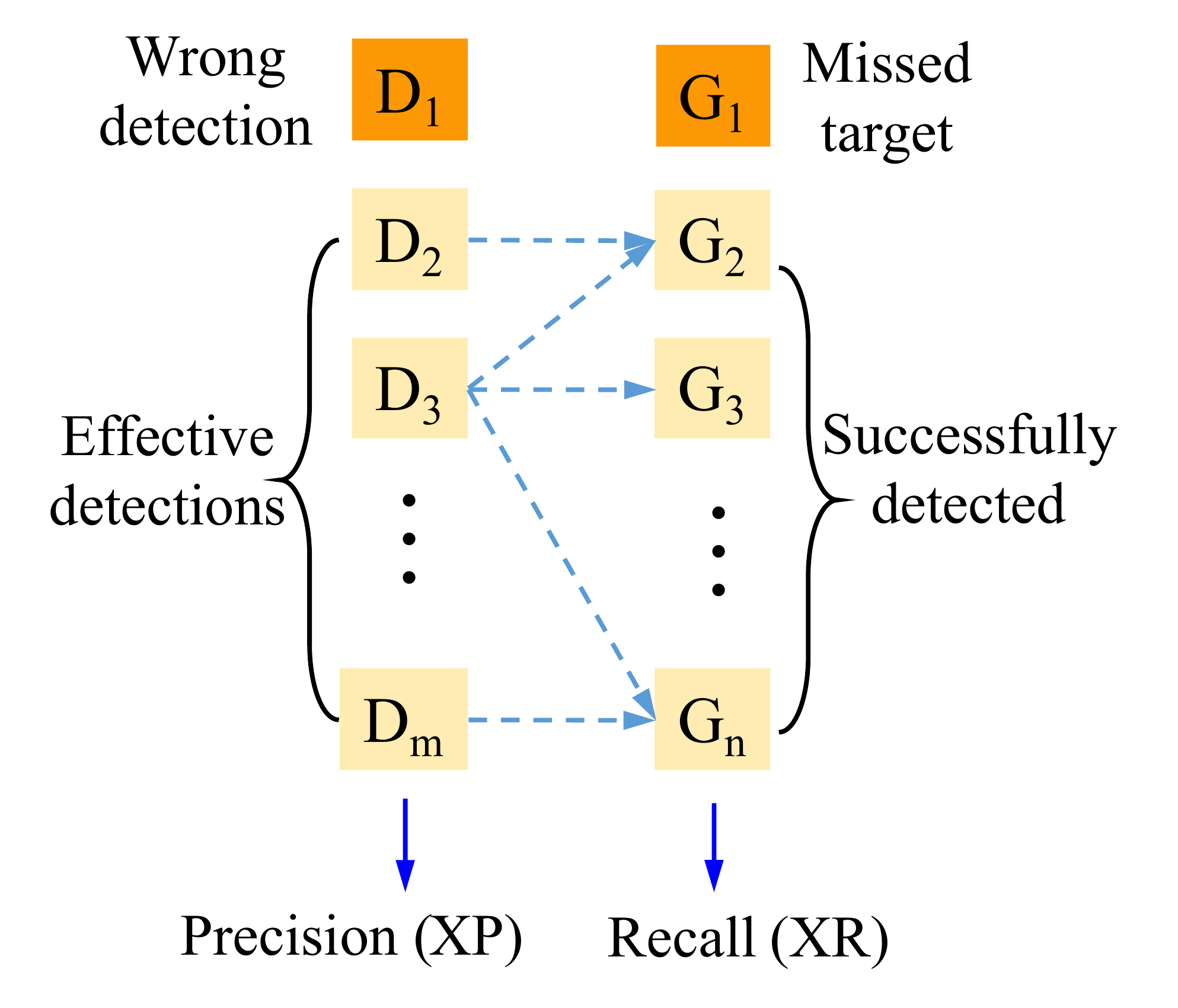}
    }
\caption{The covering box matching in CovEval for solving the \textit{scale inconsistency problem} and \textit{non-correspondence problem}.
In (a), CAr normalized the scale of boxes so boxes in different sizes can get good match.
In (b), the repetitive detection can be matched by the multi-matching process.}
\label{fig:Covering}
\end{figure}

Compared with IoU in Eq.~(\ref{fun:IoU}), we replace the denominator. This is because that for boxes in different size, $S_{G \bigcup D}$ in IoU overestimates the upper bound of $S_{G \bigcap D}$, which causes that the IoU score of these two boxes can never be high.
Without considering the scale consistency, area of the smaller box is used for scale normalization, which is a better estimation for the upper bound of $S_{G \bigcap D}$.
Thus, boxes in different sizes can get match after this normalization.
Fig.~\ref{fig:normed-overlap} illustrates this scale normalization of overlap.
By this way, we address the \textit{scale inconsistency problem}.

In fact, the overlap condition is relaxed (CAr is always greater than IoU at the same situation).
We present the following truths to ensure the effectiveness of the new overlap.
1) CAr preserves the range of $[0,1]$ which is same as IoU.
2) CAr keeps sensitivity to inaccurate positioning.
By example, for two boxes with same size, their CAr decreases to 0 together with IoU when they are getting separated.

\subsection{Covering multi-matching strategy}
\label{CAr Matrix}
Based on the CAr, we construct CAr matrixes (TABLE~\ref{tab:CArMatrix}) for box multi-matching.
In the $m\times n$ matrix for a single picture, $m$ detected boxes $\{D_i\}$ and $n$ GT boxes $\{G_j\}$ are listed.
Element at $(i,j)$ in the matrix is CAr of $D_i$ and $G_j$.
Thus, elements in the $i^{\textup{th}}$ row represent the overlaps between $D_i$ and every GT box.
It is similar in columns for every GT box.

\begin{table}[h]
\caption{The CAr Matrix.
($CAr_{i,j}$ is a shorthand of $CAr(D_i,G_j)$)}
\begin{center}
\begin{tabular}{|c|cccc|}
\hline
$CArs$ & $G_1$ &$G_2$ & $\cdots$ &$G_n$ \\
\hline
$D_1$ & $CAr_{1,1}$ & $CAr_{1,2}$ & $\cdots$ & $CAr_{1,n}$ \\
$D_2$ & $CAr_{2,1}$ & $CAr_{2,2}$ & $\cdots$ & $CAr_{2,n}$\\
$\vdots$ & $\vdots$ & $\vdots$ & $\ddots$ & $\vdots$ \\
$D_m$ & $CAr_{m,1}$ & $CAr_{m,2}$ & $\cdots$ & $CAr_{m,n}$\\
\hline
\end{tabular}
\end{center}
\label{tab:CArMatrix}
\end{table}

In practice, a detected box is valid when there is at least one GT box corresponding to it.
By counting the valid detected boxes from CAr matrix in rows, the \textit{Extended Precision} (XP) is defined as:
\begin{align}
XP=\frac{K_p}{m}
\end{align}
$K_p$ is the number of the valid detected boxes.

Analogously, a GT box for validation is successfully detected by the model when there is at least one matched detected box.
By counting the detected GT boxes from CAr matrix in columns, the \textit{Extended Recall} (XR) is defined as:
\begin{align}
XR=\frac{K_r}{n}
\end{align}
$K_r$ is the number of the detected GT boxes.
Our implementation to calculate the XP is shown in Algorithm~\ref{algorithm2}.
Using a transposed CAr matrix, steps to calculate the XR are the same.

\begin{algorithm}[ht]
\label{algorithm2}
\caption{Multi-matching strategy-- XP calculating}
\KwIn{CAr Matrix $\{CAr_{i,j}|i=1,2...n;j=1,2...m\}$ for $n$ detecting boxes and $m$ GT boxes; overlap threshold $CAr_{th}$}
\KwOut{$XP$ for one image}

$Count = zeros(n)$\;

\For{$i=1;i \leq n$}
{
    \If{$max\{CAr_{i,j}|j=1,2...m\} \geq CAr_{th}$}
    {
        set $Count_i = 1$\;
    }
}
$XP = mean(\{Count_i\})$\;
return $XP$\;
\end{algorithm}

The XP and XR are for single category in one image.
Sequentially,  \textit{Average XR} and \textit{Average XP} (AXR/AXP) can be obtained for a single class by averaging XPs/XRs of all test images.
For multi-class detection, \textit{mean AXR} and \textit{mean AXP} (mAXR/mAXP) can be obtained by averaging all AXRs/AXPs.
These steps are similar with the mAP calculation.

Compared to the mAP method, the proposed matching strategy gets rid of the one-to-one corresponding restriction.
In case of cracks, boxes should be matched by groups instead of individual ones.
In another word, we let boxes in two groups match freely without one-to-one condition.
After the matching, boxes are valid when there is at least one matched partner.
For other mismatched boxes, they are judged as false-alarm (for detected boxes) or missed targets (for GT boxes).

As the extension of \textit{Recall} and \textit{Precision}, the proposed XR and XP inherit their strong significance in practices.
\textit{Recall} represents percentage of detected crack in all targets.
\textit{Precision} denotes percentage of correct ones among all detections.
Thus, high XR ensures the reliability of automatic detectors that few cracks are missed in the detection, while XP indicates the model's ability to avoid false alarm.

\subsection{Single-value evaluation}

\begin{figure}[tb]
\centering
\vspace{-5pt}
\includegraphics[width=0.8\linewidth]{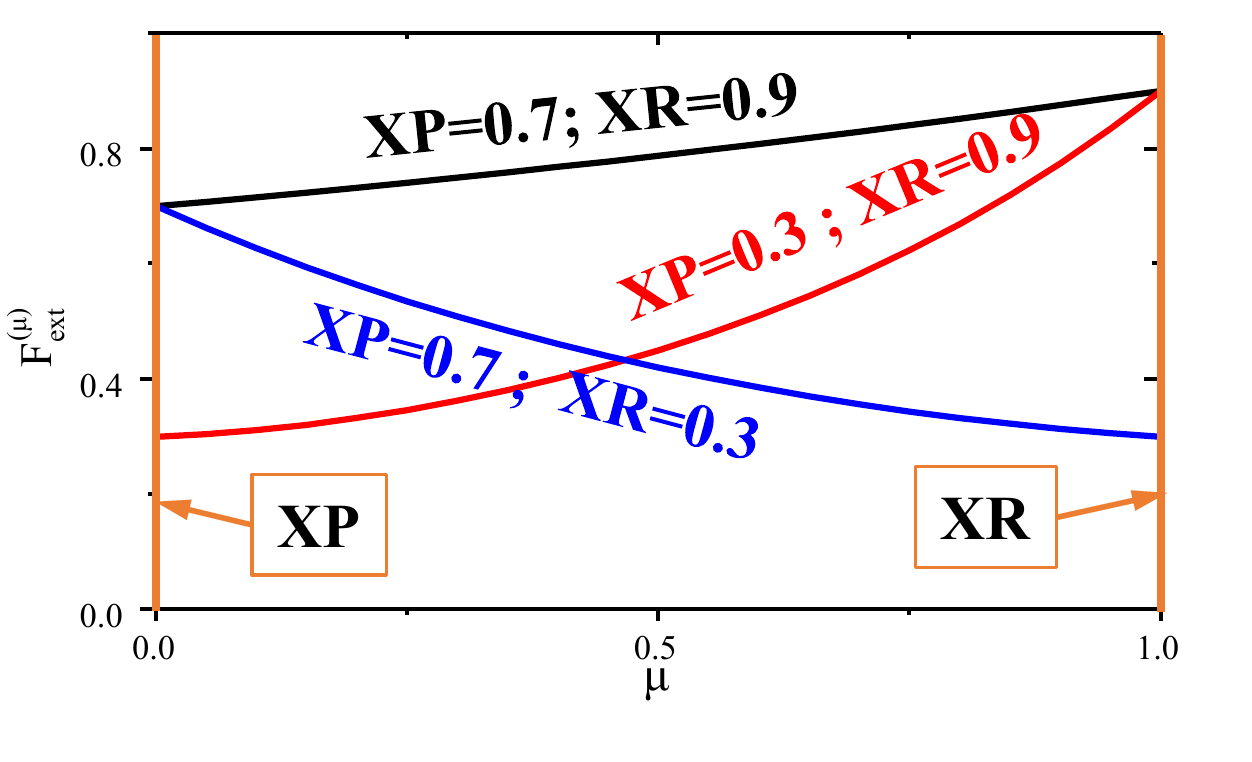}
   \vspace{-8pt}
   \caption{$F_{ext}^{(\mu)}$ curves with $\mu$ in three hypothetical cases which have constant XR and XP. Since all the curves are convex, the $F_{ext}^{(\mu)}$ tend to be closer to the smaller one of XR and XP.}
\label{fig:trade-off}
\end{figure}

Comprehensive single score for evaluation is required sometimes, especially in quantitative researches.
F-score is a conventional combination of $Recall$ and $Precision$:
$F_1=\frac{2PR}{P+R}$~\cite{Goutte2005A}.
By replacing $P$ and $R$ with $XP$ and $XR$, \textit{Extended F-score} ($F_{ext}$) can be defined as:
\begin{align}
F_{ext}=\frac{2XP \times XR}{XP+XR}
\end{align}
The $F_{ext}$ treats $Recall$ and $Precision$ equally, which is not always proper in practices. To adapt to various scenarios, a trade-off factor $\mu \in [0,1]$ is introduced to represent the bias:
\begin{align}
F_{ext}^{(\mu)}=\frac{XP^{2(1-\mu)} \times XR^{2\mu}}{(1-\mu) XP+ \mu XR}
\end{align}
This definition ensures that $F_{ext}^{(0)}=XP$, $F_{ext}^{(1)}=XR$ and $F_{ext}^{(0.5)}=F_{ext}$.
Fig.~\ref{fig:trade-off} shows how $\mu$ effects $F_{ext}^{(\mu)}$ with constant XR and XP by examples.

For various scenarios, several standard values of $\mu$ are suggested in TABLE \ref{tab:suggested-standard}.
Note that it is strongly deprecated to set $\mu$ as 0 or 1, because this may easily lead the model to perform extremely. For example, the model may mark all or none of the areas, while it can still get high score.
In crack detection, all the cracks must be detected and repaired to avoid disastrous accidents.
Thus, XR takes the dominant in crack detection.
In rest part of this paper, we mainly discuss $F_{ext}^{(0.5)}$ (the same as F-score) and $F_{ext}^{(0.8)}$ (suggested in most defect detecting tasks).

\begin{table}[h]
\vspace{-5pt}
\caption{Guidance for $\mu$ in different scenarios.}
\begin{center}
\begin{tabular}{p{0.2\linewidth}p{0.7\linewidth}}
\toprule
\makecell{Suggested $\mu$} & \makecell{Scenario \& examples} \\
\specialrule{0.5pt}{1pt}{1.5pt}
\makecell{0.05} & \textbf{Strongly avoid false alarm} \\
 & Software trigger of actuators, e.g. automatic filters on production lines \\
\specialrule{0pt}{1.5pt}{1.5pt}
\makecell{0.5} & \textbf{Balanced comprehensive scenario} \\
 & Infrastructure maintenance, e.g. road, small bridges, ceramic tiles \\
\specialrule{0pt}{1.5pt}{1.5pt}
\makecell{0.8*} & \textbf{Avoid missing the targets} \\
 & General safety confirmation of structures, e.g. large \& medium bridges, structures in rail traffic \\
\specialrule{0pt}{1.5pt}{1.5pt}
\makecell{0.95} & \textbf{Strongly avoid missing the targets} \\
 & Safety confirmation of sophisticated equipments, e.g. structures in aviation, super tall buildings \\
\bottomrule
\end{tabular}
\end{center}
* suggested in most defect detecting tasks.
\label{tab:suggested-standard}
\vspace{-10pt}
\end{table}
\begin{table*}[t]
\vspace{-9pt}
\caption{The scores of the generic models when using different evaluation standard.*}
\vspace{-8pt}
\begin{center}
\setlength{\tabcolsep}{3.5pt}
\begin{tabular}{cccccccccccccccccc}
\toprule
& \multicolumn{2}{c}{\textbf{Models}} & & \multicolumn{5}{c}{\textbf{Scores in general object detection}} & & \multicolumn{5}{c}{\textbf{Scores in crack detection}} & & \multirow{2}*{\makecell{\specialrule{0pt}{3pt}{0pt} \textbf{inference} \\ \textbf{time} (ms)}} &\\
\specialrule{0pt}{1.5pt}{0pt}
\cline{2-3} \cline{5-9} \cline{11-15}
\specialrule{0pt}{1.5pt}{0pt}
& framework & backbone & & \textbf{mAP} & $\bm{F^{(0.5)}_{ext}}$$^\ddag$ & $\bm{F^{(0.8)}_{ext}}$$^\ddag$ & mAXR$^\dag$ & mAXP$^\dag$ & &  \textbf{mAP} & $\bm{F^{(0.5)}_{ext}}$$^\ddag$ & $\bm{F^{(0.8)}_{ext}}$$^\ddag$ & mAXR$^\dag$ & mAXP$^\dag$ & &\\
\specialrule{0.8pt}{2.5pt}{2.5pt}
& \multirow{3}*{faster R-CNN} & VGG-16 & & 68.4 & 85.9 & 80.4 & 77.4 & 97.9 & & 33.2 & 89.4 & 88.5 & 87.9 & 90.9 & & $\sim$100 \\
& & ResNet-50 & & 72.2 & 89.6 & 85.3 & 82.7 & 97.8 & & 30.2 & 88.8 & 87.8 & 87.1 & 90.6 & & $\sim$100 \\
& & Mobilenetv1 & & 59.0 & 84.7 & 78.6 & 75.0 & 97.4 & & 20.5 & 88.3 & 88.9 & 89.3 & 87.3 & & $\sim$50 \\
\specialrule{0pt}{1.5pt}{1.5pt}
& \multirow{2}*{SSD} & VGG-16 & & 65.8 & 77.2 & 68.0 & 63.3 & 99.0 & & 33.1 & 83.3 & 82.9 & 82.7 & 83.9 & & $\sim$50\\
& & Mobilenetv1 & & 60.2 & 74.5 & 64.6 & 59.7 & 98.9 & & 24.7 & 79.1 & 81.8 & 83.8 & 74.9 & & $\sim$45\\
\specialrule{0pt}{1.5pt}{1.5pt}
& \multirow{2}*{YOLO~\makecell{v3 \\ v4}} & Darknet-53 & & 72.1 & 79.4 & 70.7 & 66.2 & 99.3 & & 37.0 & 88.7 & 85.1 & 85.8 & 95.5 & & $\sim$90 &\\
& & CSPDarkNet-53 & & 79.9 & 85.7 & 79.2 & 75.5 & 99.2 & & 24.0 & 86.1 & 88.4 & 90.0 & 82.5 & & $\sim$110 &\\
\bottomrule
\end{tabular}
\end{center}
\label{tab:mAP}
* All the models are run by ourselves. Scores in general object detection are the mean of 20 general categories. Settings of the models are unchanged in the general detection and the crack detection.

$^\ddag$ and $^\dag$ denote indexes proposed in CovEval, while $^\ddag$ denotes the comprehensive indexes same as the mAP.
\vspace{-5pt}
\end{table*}
\section{Validation of the CovEval standard}
\label{Experiment}

The experiment includes two parts.
1) It is validated that the proposed CovEval corrects the underestimation of mAP for the crack detectors.
2) A case study to train and optimize the model for crack detection is conducted via CovEval.
Prior to the experiments, the data set, architecture and implementation are illustrated.

\subsection{Data set, architecture and implementation}

Images containing typical cracks are collected from routine civil construction such as the concrete structure, tiled pavement, asphalt road, marble tile, etc.
They constitute a dataset named \textit{CrackSet} for crack detection. 320 of the collected 400 images are used for training and other 80 images are used for test.
All the images are horizontally and vertically flipped for data augmentation.
Besides, the \textit{pascal voc 2007}\footnote{A benchmark data set with 20 categories for \textit{object detection}.}~\cite{Everingham2006PASCALvoc} is included in our experiments for general object detection compared to the crack (Rf objects).

We adopt three popular generic frameworks, faster R-CNN~\cite{Ren2015Faster}, SSD~\cite{Liu2016SSD} and YOLO (v3 and v4) ~\cite{redmon2018yolov3,Bochkovskiy2020YOLOv4OS} as the methods for object detection in experiments.
All the settings in detail of the frameworks follow their original papers.

In evaluation, the overlap threshold is set as 0.55 for both the IoU in mAP and the CAr in proposed CovEval standard.
The confidence threshold which is specially required in CovEval is set as 0.5 to determine valid detected boxes.
Further, we use one NVIDIA TITAN Xp GPU for computation and \textit{Tensorflow}~\cite{Abadi2016TensorFlow} as the \textit{deep learning} framework.

\subsection{Results and Discussion}
\label{validation}

To validate the effectiveness of proposed CovEval, we prove two ideas in this part:

\begin{enumerate}
\item With CovEval standard, the score of crack detector increases and it corresponds well with the visual performance.
\item For general object detection, CovEval does not change greatly compared to the mAP.
\end{enumerate}

\begin{figure}[t]
\vspace{-8pt}
\centering
    \subfigure[Object categories on the score bars of AP (left) and $F^{(0.8)}_{ext}$ (right).]
        {
        \label{fig:Validation}
        \includegraphics[width=1\linewidth]{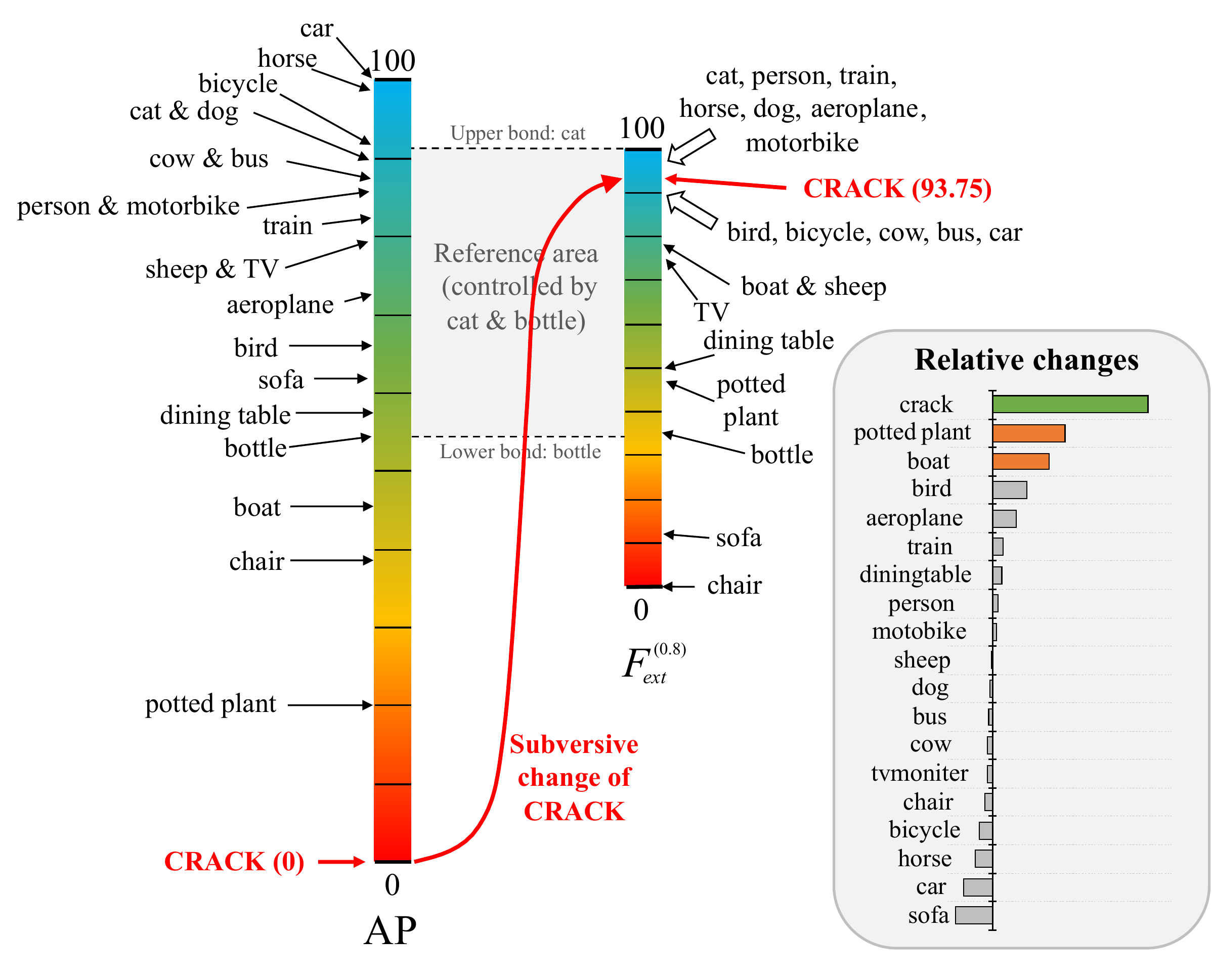}
        }
    \subfigure[Demo of potted plants]
        {
        \label{fig:pottedplant}
        \includegraphics[width=0.46\columnwidth]{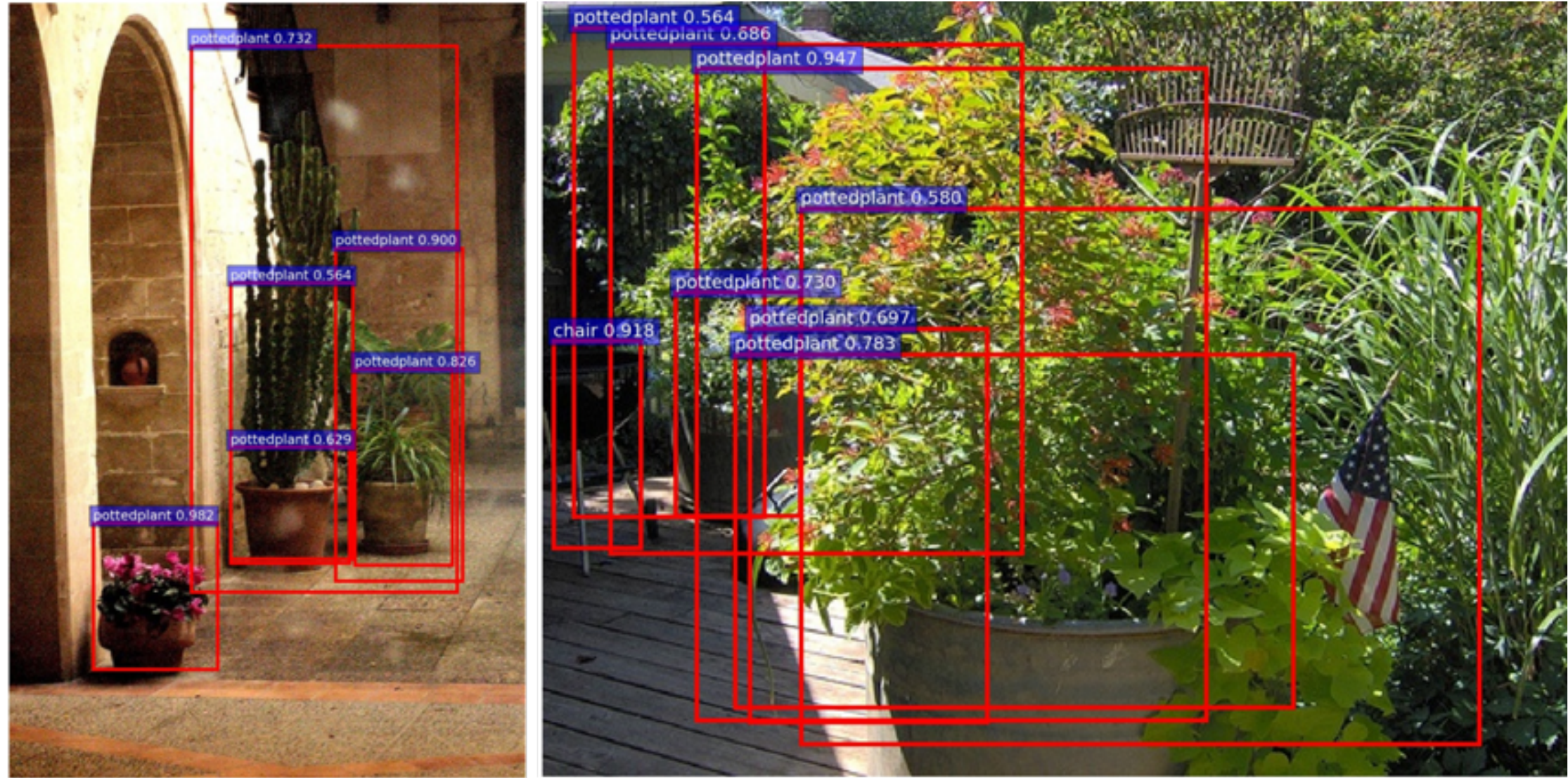}
        }
    \subfigure[Demo of boats]
        {
        \label{fig:boat}
        \includegraphics[width=0.46\columnwidth]{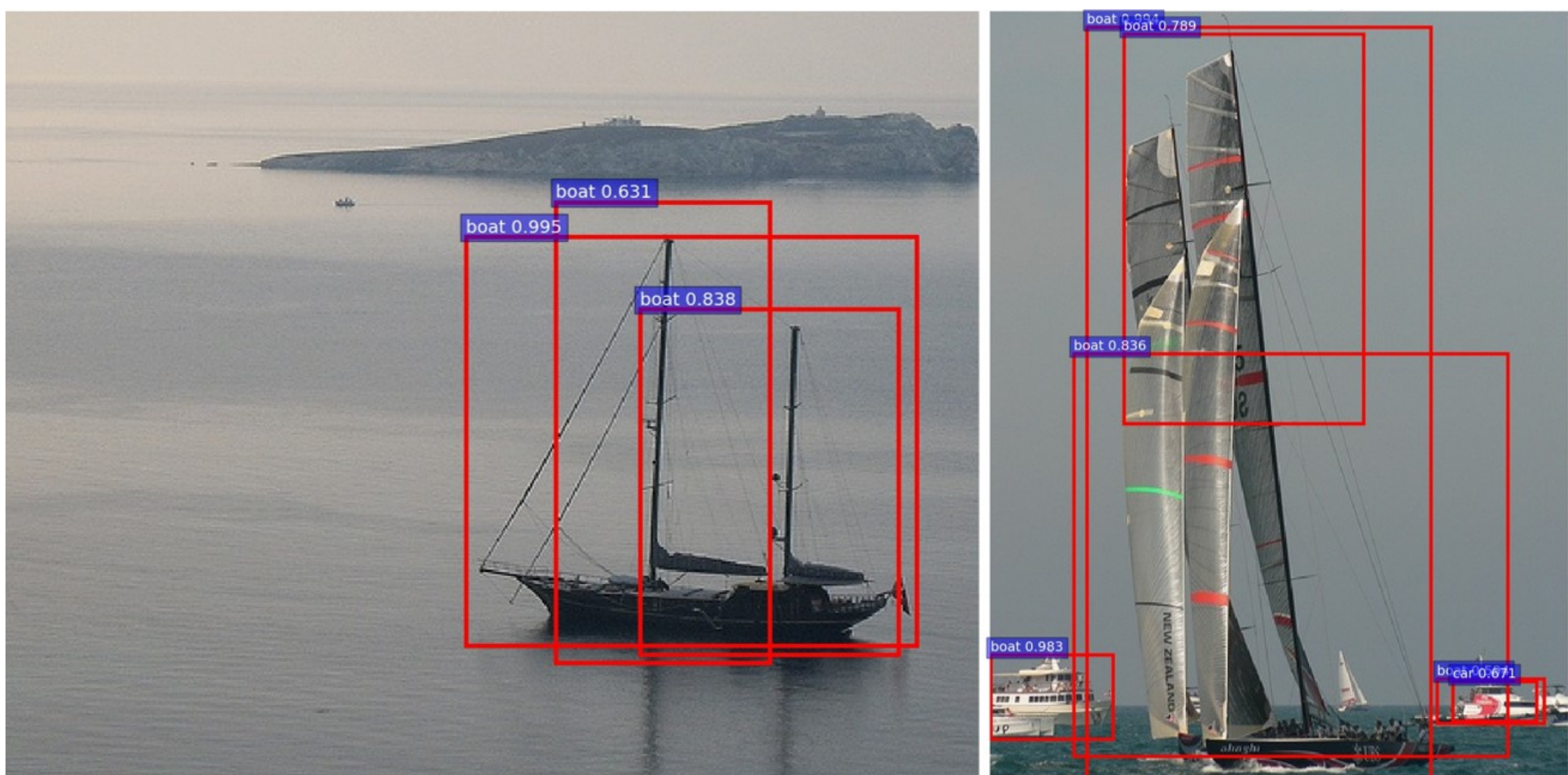}
        }
\caption{Ranked categories by the normalized scores. Identical models are scored by the mAP and the $F^{(0.8)}_{ext}$ in CovEval respectively. Taking the bottle and cat as reference boundary, the normalized scores are compared relatively.
Histogram in (a) compares the relative changes of categories.
For general categories which are more impacted by CovEval, the demos are visualized in (b) and (c). The boxes show a similar mode with the crack detection, which could be the reason for being sensitive to CovEval.}
\vspace{-8pt}
\end{figure}

The evaluation standard proposed is for methods for \textit{object detection}, which includes a considerable number of different models.
Seven different models based on three popular generic frameworks are trained to detect crack and general objects.
These models are scored by CovEval and mAP respectively, and the result shows the improvement of CovEval in crack detection.
We highlight the truths in TABLE~\ref{tab:mAP} here for supporting the two ideas above:
\begin{enumerate}
\item The scores of mAP for models in crack detection are really poor (7 models from 20.5 to 37.0) compared to the general detection (7 models from 59.0 to 79.9).
\item The score of identical model in crack detection from CovEval (e.g. the $F^{(0.5)}_{ext}$ and $F^{(0.8)}_{ext}$) is competitive. It is slightly higher than the scores in general detection.
\item Generally, the scores for models from CovEval in general detection follow the trend of mAP.
\end{enumerate}
Demo images in Fig.~\ref{fig:motivation} are from the faster R-CNN model with VGG-16 backbone.
Comparing the demos with the scores, CovEval reflects the visual performance of the model well and it corrects the underestimation of mAP standard.

One of the listed models, the faster R-CNN framework with VGG-16 backbone, is selected for in-depth investigation about the two standards.
The scores for the 21 categories (includes 20 general categories in \textit{pascal voc 2007} and the crack) are sorted in Fig.~\ref{fig:Validation} by the normalized AP and normalized $F^{(0.8)}_{ext}$ respectively.
According to the histogram in Fig.~\ref{fig:Validation}, the scores of crack obtains a subversive change (from 21$^{\textup{st}}$ to 8$^{\textup{th}}$), while the general categories do not change much on the common relative axis between the two standards.
Note that the two categories, the potted plants and the boats, changes greatly in score among all general categories.
The reason is shown in Fig.~\ref{fig:pottedplant} and~\ref{fig:boat}.
The kinds of objects are often marked by boxes which only include partial features of the whole object.
This phenomenon is similar to crack (Rf object), so their scores are improved.
But it is worth noting that these general objects are essentially different from cracks theoretically.
The features in any subpart of the crack are actually complete for the whole object (It is theoretically proved in Section~\ref{theory}), but the local features of general objects in these boxes are only sufficient sometimes for the object inference.

\section{Optimize the practical model via CovEval}
\label{fine-tuning}

Applying the proposed CovEval standard, a practical model for automatic crack detection based on \textit{deep learning} is trained and optimized based on the faster R-CNN framework with VGG-16 backbone~\cite{Ren2015Faster}.
Considering the requirements in industry inspection, we adopt $F^{(0.8)}_{ext}$ and AXR as the dominant indexes in this study.

The optimization of the faster R-CNN framework focuses on 3 parameters: 1) the anchor scales and ratios in \textit{Region Proposal Network} (RPN), 2) the batch size of proposals in \textit{Region of Interest set} (RoIs), 3) the $learning$ $rate$ ($lr$) schedule.
The anchors in RPN decide shape of the default boxes in these region-based models, which are proved influential to the final performance~\cite{yang2018metaanchor}. Besides, the batch size and the $lr$ schedule are important in model training.
Models trained with smaller batch size are more likely to perform better and models trained with bigger $lr$ converge faster, but they both makes the results not so stable.
In fact, it is ordinary to conduct this study, however, it is not implementable without the fair standard CovEval proposed.

\begin{figure}[tbp]
\centering
\includegraphics[width=\linewidth]{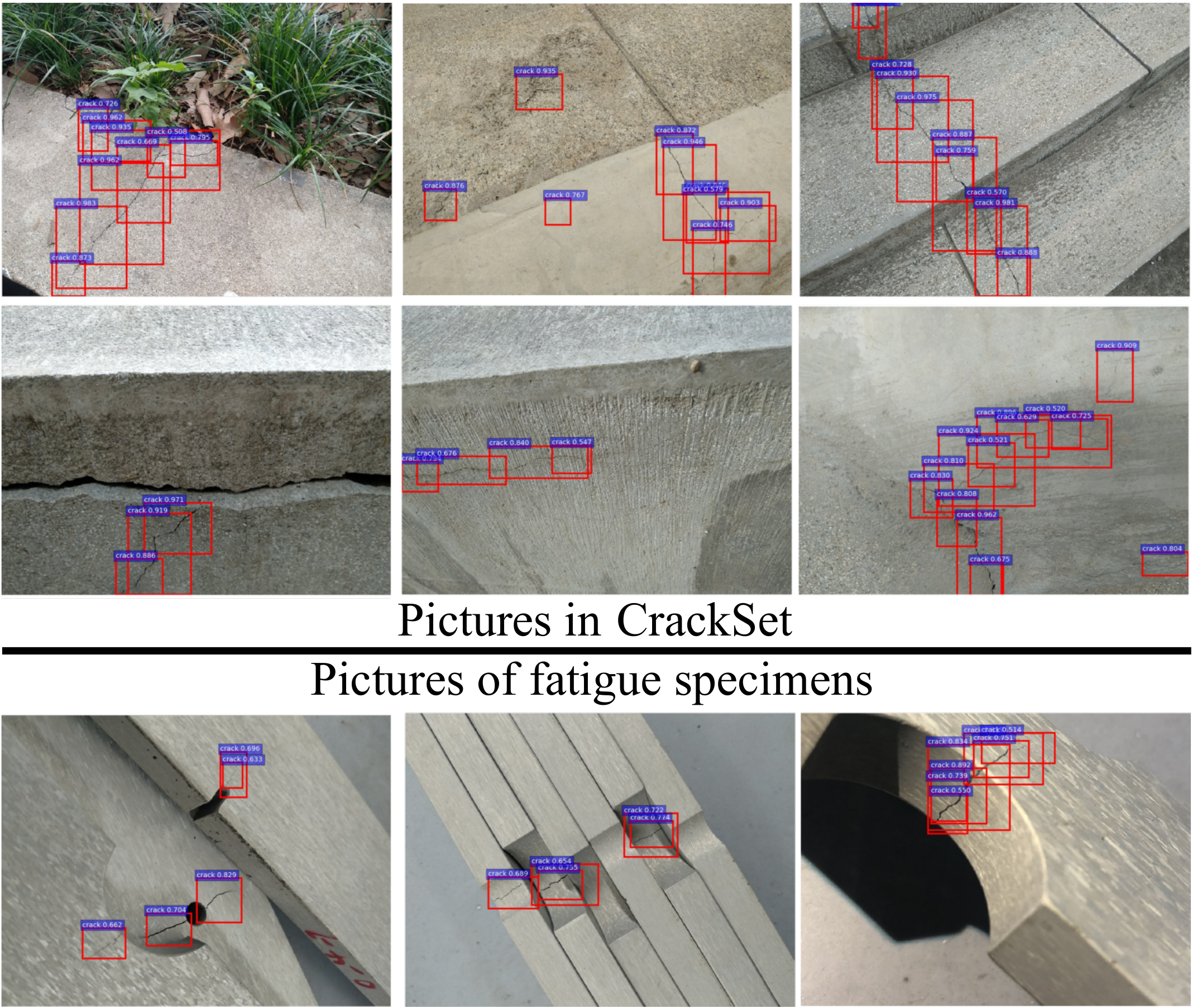}
   \caption{Images with red boxes of detected cracks. Pictures above the horizontal line are from test set of \textit{CrackSet}. The rest pictures not included in \textit{CrackSet} are cracks on fatigue specimens. It shows a different scenario for application.}
\label{fig:results}
\end{figure}

After experiments, the optimal parameters are found as follow.
1) The anchor scales and ratios are [4,8,16] and [0.2,1,5] respectively.
2) The batch size of proposals in RoIs is 64;
3) With \textit{Stochastic Gradient Descent} (SGD) as the optimizer, the $lr$ adopts a two-stage schedule. It is set as $5 \times 10^{-4}$ at first, then divided by 10 at the 3000$^{\textup{th}}$ iteration. $F^{(0.8)}_{ext}$ and AXR of the best model achieve 91.64 and 93.6 respectively, which are much higher than the scores before the optimization (the previous performance is $F^{(0.8)}_{ext}$ at 88.2 and AXR at 87.9 in TABLE~\ref{tab:mAP}). After changing the confidence threshold from 0.5 to 0.1, $F^{(0.8)}_{ext}$ achieves 92.2 while the AXR is 95.8.
Fig.~\ref{fig:results} presents the demo images from our best model, and the performance of the model is pretty good.
Besides, cracks on other images of fatigue specimens in the laboratory are also detected.
These images show that the trained model is really robust to different materials and environments.

\section{Conclusion}
\label{Conclusions}

In this work, the following speculations are proved theoretically and experimentally.
1) The models for automatic crack detection are strongly underestimated by traditional mAP standard.
2) The random fractal of crack is denoted as the origin of the mAP¡¯s failure, because the strict box matching in mAP calculation is unreasonable for evaluating the crack detectors.
Taking advantage of the \textit{covering} strategy in box matching, a new evaluation standard CovEval is proposed, which addresses problems for evaluating the crack detectors.
Experiments show that the underestimation for the \textit{object detection} models in crack detection is corrected by CovEval.
Applying CovEval, the generic models for \textit{object detection} based on \textit{deep learning} show strong ability in crack detection that the best model achieves recall at 95.8.

As a fair standard, CovEval addresses the issues in evaluating the methods for crack detection.
We hope it can promote studies and applications about the methods for crack detection in industry.

\appendices
\section{Proof of All-scale Equivalence of the Fractal}
\label{proof1}

\begin{proof}
The curve composed of line segments can be determined by a point sequence $\{P_i\}$, e.g. a curve $C$ includes $M$ segments and $M+1$ points can be represented as $C(\{P_i|i=0,1,...,M\})$ or $C(P_0,P_M)$.

Following Eq.~(\ref{iteration}, \ref{fractal_definition}), $G$ points (for curves in Fig.~\ref{regular_fractal} and~\ref{random_fractal}, $G=1$) are inserted in every interval by the generating method $\mathscr{F}$ in iterations. Therefore, all inserted points in $C_{f}$ can be indexed by its inserting iteration $n$ and its group order $k$ (illustrated in Fig.~\ref{fig:indexed}). We can get the point sequence $\{P(n,k)\}$ for $C_{f}$. There are implicit constraints that
$k \le G \cdot (G+1)^{n-1}$,
and $\forall n \le N_0$, we have $P(n,k) \in C^{(N_0)}$ where $C^{(N_0)}$ is the intermediate curve after $N_0$ iterations.

\begin{figure}[ht]
\centering
\includegraphics[width=0.9\linewidth]{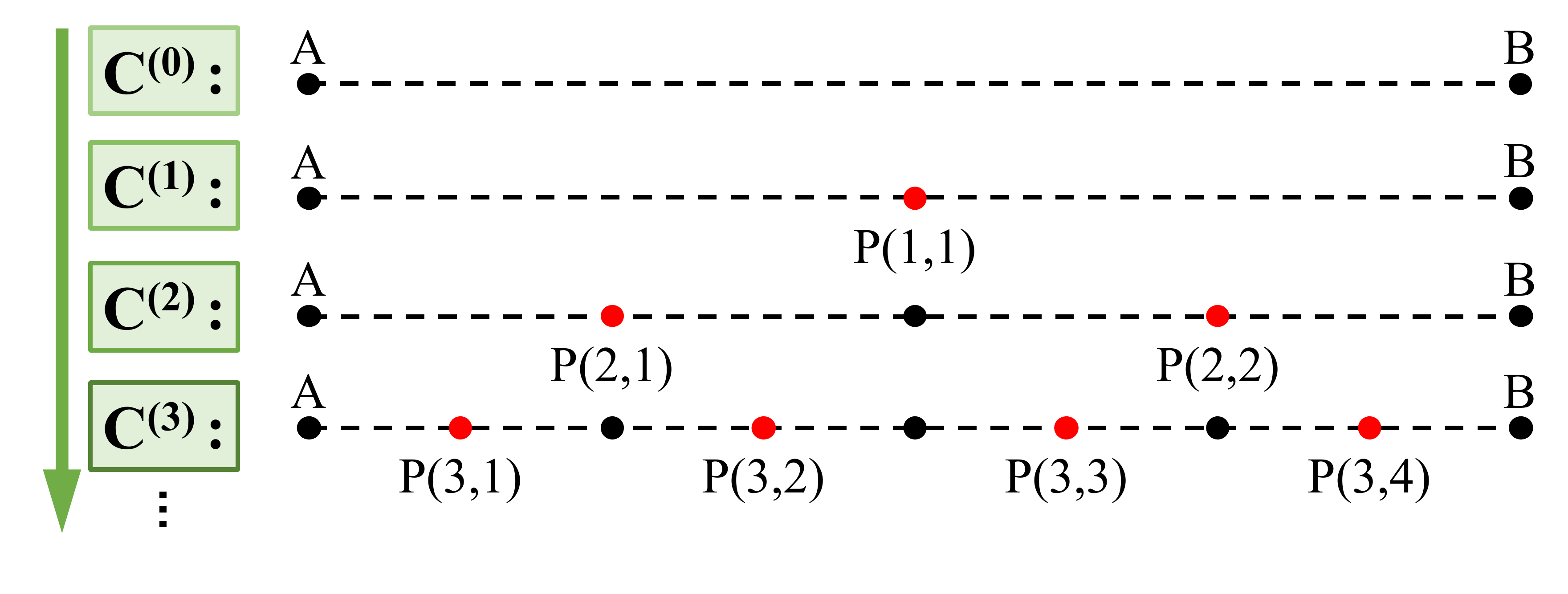}
\vspace{-8pt}
\caption{Indexed points in fractal curves. Points on the dotted lines only represent their topological relation in the fractal curves.}
\label{fig:indexed}
\end{figure}

Define a topological order $T$ for every indexed point to sort the points in curves:
\begin{align}
T_{P(n,k)}=\frac{k + \lfloor (k-1)/G \rfloor}{(G+1)^{n}}
\end{align}
$P_\alpha(n_\alpha,k_\alpha)$ is always in front of $P_\beta(n_\beta,k_\beta)$ if $T_{P_\alpha}<T_{P_\beta}$.

On this basis, for $C_s(P_a,P_b) \subseteq C_{f}(P_0,P_\infty)$,

$\because$ $C_s$ is not infinitesimal

$\therefore \exists$ a positive integer $N$ that have
\begin{align}
\begin{split}
T_{P_b}-T_{P_a} &= \frac{k_a + \lfloor (k_a-1)/G \rfloor}{(G+1)^{n_a}} - \frac{k_b+ \lfloor (k_b-1)/G \rfloor}{(G+1)^{n_b}} \\
&>\frac{2}{(G+1)^N} = 2T_{P(N,1)}
\end{split}
\end{align}

Note that difference (distance) of the topological order is the same for adjacent points in $C^{(N)}$, and they are all equal to $T_{P(N,1)}$. Thus, we can find at least two points, $P_\alpha(n_\alpha,k_\alpha)$ and $P_\beta(n_\beta,k_\beta)$, in $C_s(P_a,P_b)$ that have
\begin{equation}
\left\{
    \begin{array}{lr}
    n_\alpha, n_\beta \le N \\
    \left| T_{P_\alpha(n_\alpha,k_\alpha)} - T_{P_\beta(n_\beta,k_\beta)} \right| = T_{P(N,1)} \\
    \end{array}
\right.
\end{equation}
This condition means the $P_\alpha$ and $P_\beta$ are adjacent points in $C^{(N)}$, i.e. $\overline{P_\alpha P_\beta} \in C^{(N)}$.
Rewrite the $\overline{P_\alpha P_\beta}$ as $\overline{A'B'}$ and regard it as the original segment for another fractal curve.
All other points between $P_\alpha$ and $P_\beta$ are inserted by same method $\mathscr{F}$ iteratively.
Writing the subcurve of $C_s$ between $P_\alpha$ and $P_\alpha$ as $C_1(P_\alpha,P_\beta)$, we have
\begin{align}
\begin{split}
C_1(P_\alpha,P_\beta)&=\lim_{n \to \infty} \mathscr{F}^{(n-N)}(\overline{P_\alpha P_\beta}) \\
&\sim \lim_{n \to \infty} \mathscr{F}^{(n)}(C^{(0)})=C_{f}
\end{split}
\end{align}
The mark, '$\sim$', represents equivalent graphic feature here.

After removing $C_1$ from $C_{s}$, the rest parts $\{\hat{C}_i\}$ are also continuous subparts of $C_f$.
For example after the first round, there are two rest parts, $\hat{C}_2(P_a,P_\alpha)$ and $\hat{C}_3(P_\beta, P_b)$ ($i=2,3$, and assume that the 4 points are different). Following similar steps, subpart $C_{i}$ that has equivalent features with $C_f$ can be easily found in every remaining $\hat{C}_{i}$. This operation can be repeated until all the remaining parts vanish or become infinitesimal, which means there is no gap between $\{C_{i}\}$ and the $C_s$ is composed by $\{C_{i}\}$.

Note that for random fractal curve, the equivalence of graphic features are statistical. This completes the proof.
\end{proof}

\ifCLASSOPTIONcaptionsoff
  \newpage
\fi

{\footnotesize
\bibliographystyle{IEEEtran}
\bibliography{rfbib}
}

\end{document}